\documentclass[letterpaper, 10 pt, conference]{ieeeconf}  % Comment this line out if you need a4paper

\pdfoutput=1                                              % Use this for arxiv
\IEEEoverridecommandlockouts                              % This command is only needed if 
\usepackage{cite}
\usepackage{amsmath,amssymb,amsfonts}
\usepackage{graphicx}
\usepackage{caption}
\usepackage{subcaption}
\usepackage{textcomp}
\usepackage{xcolor}
\usepackage{float}
\usepackage{svg}
\usepackage{amsmath,amssymb}
\usepackage{hyperref}
\usepackage{verbatim}
\usepackage[normalem]{ulem}
\usepackage{todonotes}

\graphicspath{ {images/} }

\def\BibTeX{{\rm B\kern-.05em{\sc i\kern-.025em b}\kern-.08em
    T\kern-.1667em\lower.7ex\hbox{E}\kern-.125emX}}

\title{\LARGE \bf Data-driven vehicle speed detection from synthetic driving simulator images}

\author{A. Hern\'{a}ndez Mart\'{i}nez$^{1}$, J. Lorenzo D\'{i}az $^{1}$, I. Garc\'{i}a Daza$^{1}$ and D. Fern\'{a}ndez Llorca$^{1,2}$% <-this % stops a space
\thanks{$^{1}$ Computer Engineering Department, Polytechnic School, University of Alcal\'a, Madrid,  Spain. \{antonio.hernandezm, javier.lorenzod, ivan.garciad, david.fernandezl\}@uah.es \newline
$^{2}$ European Commission, Joint Research Center, Seville, Spain. david.fernandez-llorca@ec.europa.eu}%
}

\begin{document}

\maketitle
\thispagestyle{empty}
\pagestyle{empty}

\begin{abstract}
Despite all the challenges and limitations, vision-based vehicle speed detection is gaining research interest due to its great potential benefits such as cost reduction, and enhanced additional functions. As stated in a recent survey \cite{Llorca2021}, the use of learning-based approaches to address this problem is still in its infancy. One of the main difficulties is the need for a large amount of data, which must contain the input sequences and, more importantly, the output values corresponding to the actual speed of the vehicles. Data collection in this context requires a complex and costly setup to capture the images from the camera synchronized with a high precision speed sensor to generate the ground truth speed values. In this paper we explore, for the first time, the use of synthetic images generated from a driving simulator (e.g., CARLA) to address vehicle speed detection using a learning-based approach. We simulate a virtual camera placed over a stretch of road, and generate thousands of images with variability corresponding to multiple speeds, different vehicle types and colors, and lighting and weather conditions. Two different approaches to map the sequence of images to an output speed (regression) are studied, including CNN-GRU and 3D-CNN. We present preliminary results that support the high potential of this approach to address vehicle speed detection. 
\end{abstract}

\section{Introduction}
Accurate and efficient detection of vehicle speed from external systems is a well-established area of research and development, which in recent years is attracting more research attention due to the impact these systems have on road safety. Speed enforcement is a key road safety measure that directly leads to a reduction in accidents \cite{ERSO2018}. Indeed, in the vicinity of speed cameras the reduction of speeding vehicles and crashes can reach up to $35\%$ and $25\%$ respectively \cite{Wilson2010}. In addition, it has been proven that, the higher the intensity of enforcement, the greater the reduction of accidents \cite{Elvik2011}. 

The requirements for accuracy and robustness in vehicle speed measurement for speed enforcement are very demanding \cite{Llorca2016}. Thus, the most common sensors used for speed estimation are based on high-accuracy, high-cost range sensors, such as radar and laser, and, in some cases, on sensors in the pavement embedded in pairs under the road surface. The use of cameras has rarely been proposed for speed enforcement. They usually play a secondary role in capturing human-understandable scene information, allowing visual identification of vehicles and serving as evidence when the speed limit is violated.

\begin{figure}[t]
    \centering
    \includegraphics[width=\linewidth]{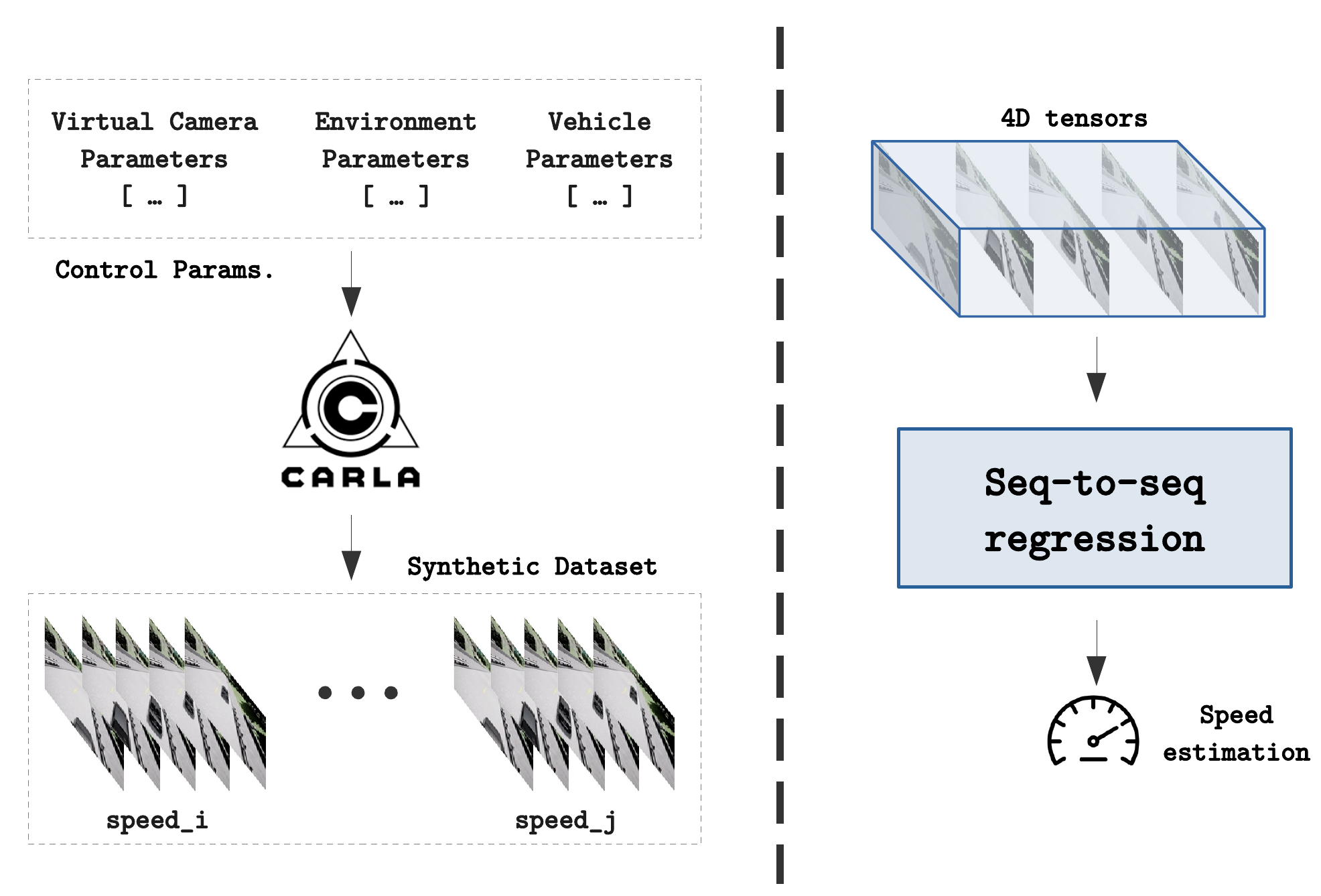}
    \caption{General overview of the data-driven vision-based vehicle speed detection approach from driving simulator images.}
    \label{fig:Intro}
    %\vspace{0mm}
\end{figure}

However, the potential benefits in terms of cost reduction and enhanced functionality, as well as recent advances in the field of computer vision have led to a significant increase in the number of works using vision as the sole mechanism for measuring vehicle speed \cite{Llorca2021}. This a challenging problem due to the discrete nature of video sensors, in which the accuracy of the representation decreases proportionally to the square of the distance, and the worsening of performance in adverse weather conditions. An additional difficulty is the limited availability of data in real environments that allow the deployment of learning-based approaches. Data collection in this domain requires a complex, and costly, setup to capture the images from the cameras synchronized with some high precision speed sensor to generate the ground truth speed values. The number of datasets with this type of information is still very limited, and this implies that the use of data-driven strategies is far from being consolidated in this application context.

In this paper, we study, for the first time, the use of synthetic sequences generated from a driving simulator to address vehicle speed detection using a learning-based approach. We present a highly realistic synthetic dataset with sequences generated from a virtual camera placed over a stretch of road, for the development and evaluation of vehicle speed detection methods. We use the open-source CARLA simulator for autonomous driving research \cite{carla2017} which allows the variation of multiple parameters, including the road layout, vehicle dynamics, different types of vehicle and colors, as well as multiple lighting and weather conditions. We then address the vehicle speed detection problem as a sequence-to-sequence regression problem \cite{Sutskever2014} using two different methods: CNN-GRU \cite{Donahue2017} and 3D-CNN \cite{Ji2013}. An overview of our approach is depicted in Fig. \ref{fig:Intro}. We present preliminary results with only one virtual camera, which support the proposed methodology and validates the usefulness of our synthetic vehicle speed detection dataset.

\section{Related work}
\label{sec:relatedwork}
As stated in a recent survey \cite{Llorca2021}, despite the fact that we can find hundreds of works focusing on vision-based vehicle speed detection, the problem is still in a moderate level of maturity. Multiple issues remain open, such as robustness in difficult lighting and weather conditions, sub-optimal system settings resulting in very high meter-to-pixel ratios, the lack of well-established datasets that allow to compare the performance of different methods, and the still small number of data-driven approaches which are well consolidated in other computer vision areas. 

In this paper, we focus on learning-based methods, so we only analyze this type of approaches in the state-of-the-art presented.  We refer to \cite{Llorca2021} for a comprehensive and detailed overview of the vehicle detection problem from non-learning based approaches, including a complete taxonomy that categorizes all the stages involved. 

\subsection{Data-driven approaches}
Concerning the estimation of vehicle speed using learning-based methods, only a few works can be found in the available literature. In \cite{Lee2019}, the authors propose the use of a CNN to estimate the mean speed of all vehicles using two consecutive images from a top view of a two-lane road segment. They propose to use two types of datasets, one using images from a long-distance video camera over a stretch of road, and the other using a synthetic set generated by a cycle-consistent adversarial network (Cycle-GAN) that converts animation images into realistic photos. Although this approach cannot be applied for accurate detection of the speed of each individual vehicle, and despite the limitations and fluctuations of the adversarial network to generate realistic images, the idea of using synthesized images to train a CNN model to estimate the average speed of traffic, which is then tested on real images, is worth mentioning. In \cite{Dong2019}, average traffic speed estimation is addressed as a video action recognition problem using 3D CNNs. They propose to concatenate RGB and optical flow images. They found that optical flow was crucial information for the vehicle speed estimation problem. However, as stated in the paper, the main limitation of the proposed model was the lack of data which could lead to overfitting. Finally, in \cite{Madhan2020} a Modular Neural Network (MNN) architecture is presented to perform joint vehicle type classification and speed detection. 

In all these cases, the input sequence corresponds to a road section with multiple lanes and vehicles, making them more suitable for traffic speed detection. The question is how to extend these approaches to perform individual vehicle speed detection. As it happens when applying video action recognition approaches for the classification of vehicle lane changes \cite{Biparva2021}, it could be necessary to generate regions of interests (ROIs) with sufficient spatio-temporal information to distinguish different speeds for different types of vehicles and scenarios. This problem
can also be mitigated by using a camera setting with a very low meter-pixel ratio focusing on a single road lane.

\subsection{Datasets}
As described in \cite{Llorca2021}, we can only find two datasets with speed ground truth available. First, we have the \emph{BrnoCompSpeed} benchmark \cite{Sochor2017}, which contains 21 sequences at a resolution of 1920 $\times$ 1080 pixels at 50fps, each sequence of 1 hour duration, including almost 21K tagged vehicles. Speed ground truth is provided using a laser-based light barrier system. The sequences were recorded in real traffic conditions in highway scenarios. The images cover between 2 and 3 lanes, in road segments of several dozens of meters. Therefore, the images have a high-medium meter-to-pixel ratio. Second, we have the \emph{UTFPR} dataset \cite{Luvizon2017} which contains up to 5 hours of sequences of images from a urban environment (speed limit at 60 km/h), 1920 $\times$ 1080 of pixel resolution at 30fps, recorded from one camera covering up to 3 lanes, and includes different weather conditions. The ground truth speeds are obtained using inductive loop detectors. The camera position has a considerable pitch, which produces a favorable medium-low meter-to-pixel ratio. 

These datasets contain valuable information for the purpose of vehicle speed detection, as they contain real data. However, they lack the versatility that is available in a simulated environment, where we can easily vary the camera parameters, vehicle types, speed range, road layout, lighting and weather conditions, etc. Given that cars are rigid objects, and thanks to the constantly improving visual realism offered by driving simulators, this is undoubtedly a favourable application context where the generation of synthetic sequences for the development and evaluation of different methods, has the greatest potential.

\section{Method}
 \label{sec:method}

This section describes, on the one hand, the process of generating the synthetic dataset and, on the other hand, the CNN-based regression models used for speed estimation. 

\subsection{Dataset construction}
\begin{figure}[ht]
    \centering
    \includegraphics[width=0.8\linewidth]{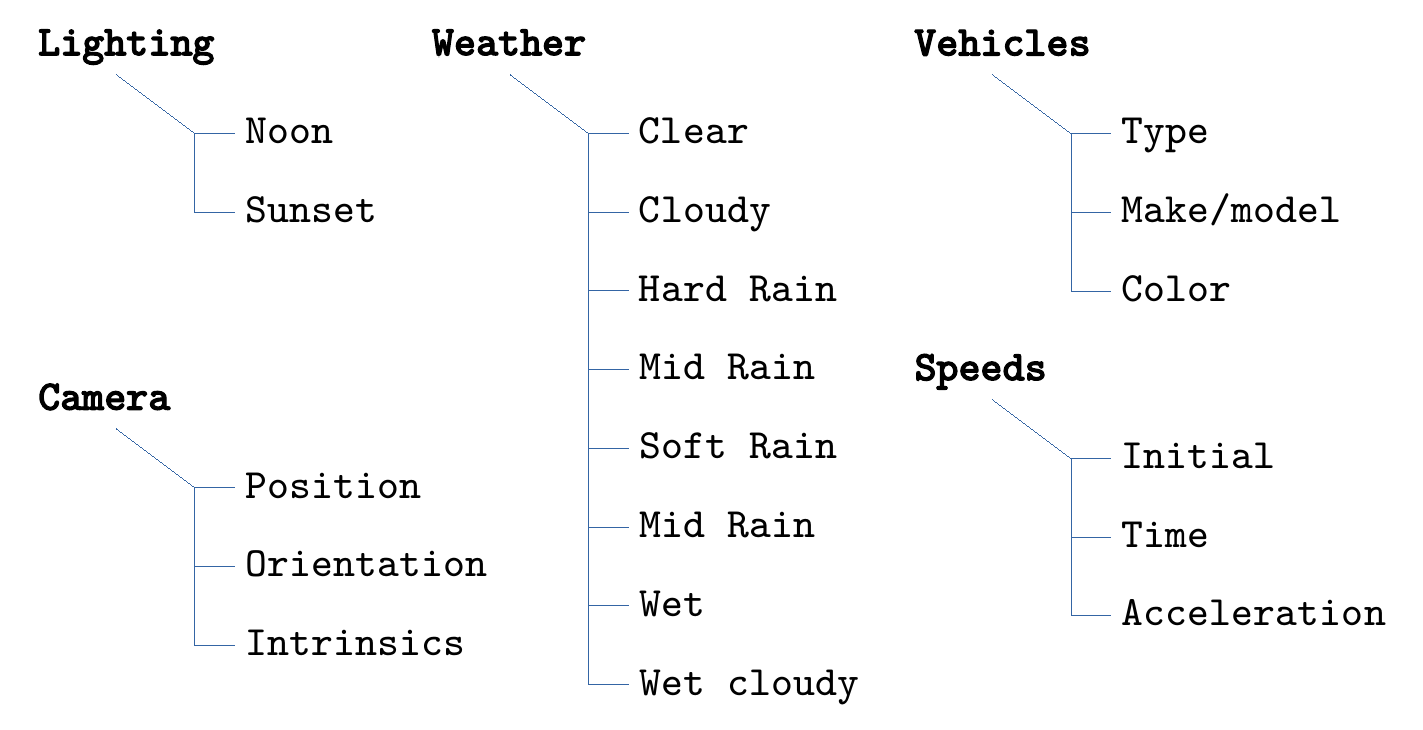}
    \caption{Simulation parameters: camera, vehicle, speeds, lighting and weather conditions.}
    \label{fig:simvars}
    %\vspace{0mm}
\end{figure}

The synthetic dataset was generated using the CARLA driving simulator \cite{carla2017} as it features a high degree of visual realism in all its scenarios, thanks to the UnrealEngine graphics engine. It allows to use a wide variety of vehicles, weather and lighting conditions, as well as the placement and orientation of one or multiple cameras. All the parameters that can be adapted to generate the synthetic vehicle speed detection dataset are depicted in Fig. \ref{fig:simvars}. 

We construct synthetic temporal sequences using one fixed camera at 80 FPS with Full HD format (1920x1080), located at a height of 3 meters, with a pitch angle of 45 degrees, and with the Z-axis parallel to the direction of travel of the vehicles. A straight road section has been selected from the map labeled as \textit{"Town01"}. Up to 27 different vehicles were used including multiple car makes and models (e.g., Audi etron, Citröen C3, Toyota Prius, etc.), trucks, motorbikes, and bikes, with different colors. 

% The vehicles used, including motorbikes, are: audi.a2, audi.etron, audi.tt, bh.crossbike, bmw.grandtourer, bmw.isetta, carlamotors.carlacola, chevrolet.impala, citroen.c3, diamondback.century, dodge\_charger.police, gazelle.omafiets, harley-davidson.low\_rider, jeep. wrangler,kawasaki.ninja, lincoln.mkz2017, mercedes-benz.coupe, mini.cooperst, mustang.mustang, nissan.micra, nissan.patrol, seat.leon, tesla.cybertruck, tesla.model3, toyota.prius, volkswagen.t2, yamaha.yzf, additional information on each vehicle used is available at \cite{carla2017}.

Multiple lighting and weather conditions have been simulated. The solar elevation angle was discretised to two positions \emph{Noon} ($75^{\circ}$) and \emph{Sunset} ($15^{\circ}$). The precipitation percentage was discretised to values of (0, 15, 30 and 60), and the precipitation deposition percentage on the surface was discretised to (0, 50, 100). Although the simulator also allows the modification of other parameters, such as cloudiness or ambient fog, they have not been taken into account in this preliminary study. 

\begin{figure}[t]
    \centering
    \includegraphics[width=0.9\linewidth]{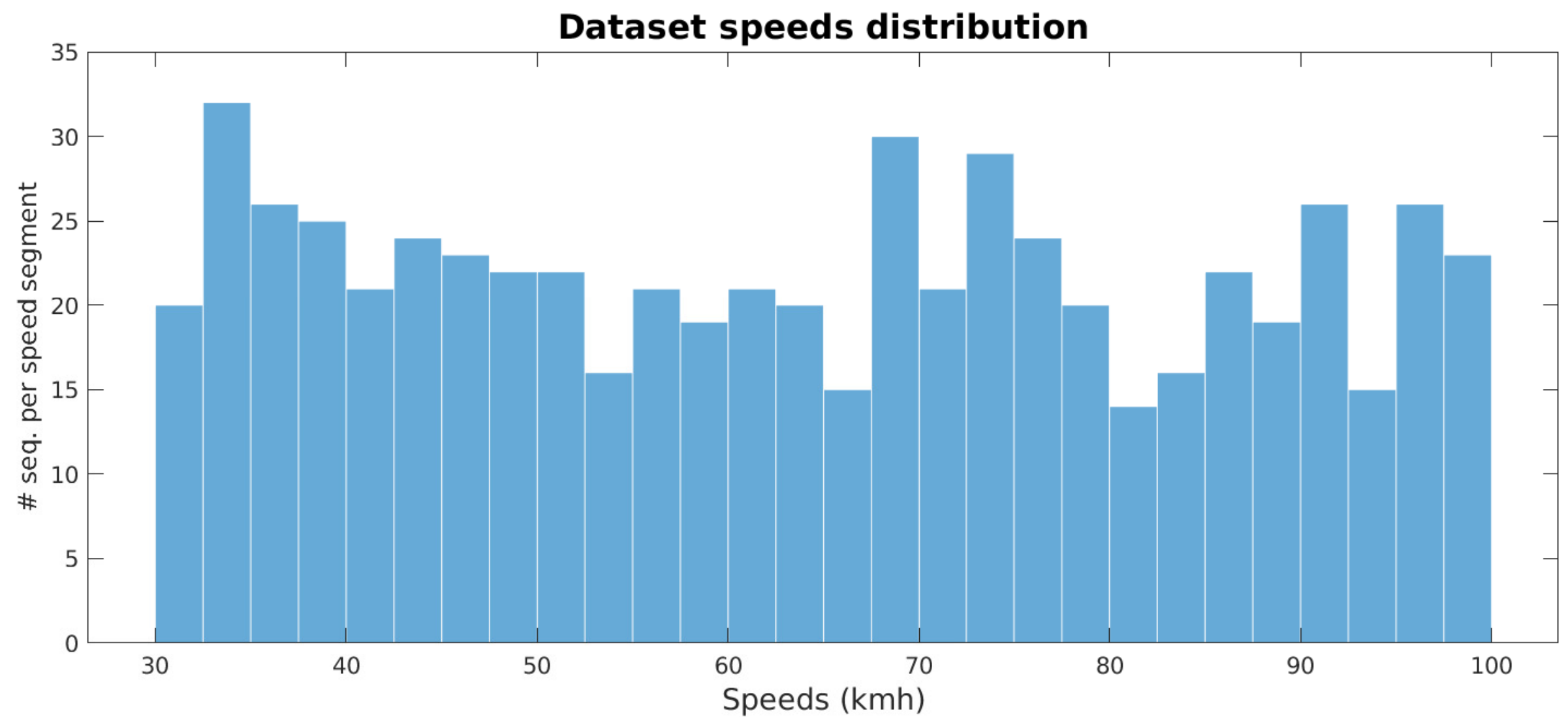}
    \caption{Distribution of speeds in the dataset.}
    \label{fig:SpeedDist}
    %\vspace{0mm}
\end{figure}

The number of images generated in each sequence will depend on the speed of the vehicle. In this case, preliminarily, the sequences are generated with constant speeds (zero acceleration). The (initial) speeds are generated randomly according to the following equation:

\begin{center}
\begin{equation} \label{eqn:rand_speed}
Speed(i) = 8.33 + X(i)\cdot19.44
\end{equation}
\end{center}
where $i$ is the episode, and $X(i)$ is a randomly selected number from a uniform distribution between 0 and 1. In this way, speeds between 8.3 and 27.7 m/s ( $\approx$ 30 and 100 km/h) are obtained. The obtained speed distribution is depicted in Fig. \ref{fig:SpeedDist}. This range of speeds has been preliminarily selected in order to validate the methodology for urban environments. For the case of highway or secondary roads, the maximum speed should be extended, as well as the camera capture FPS. 

\begin{figure}[ht]
    \centering
    \includegraphics[width=\linewidth]{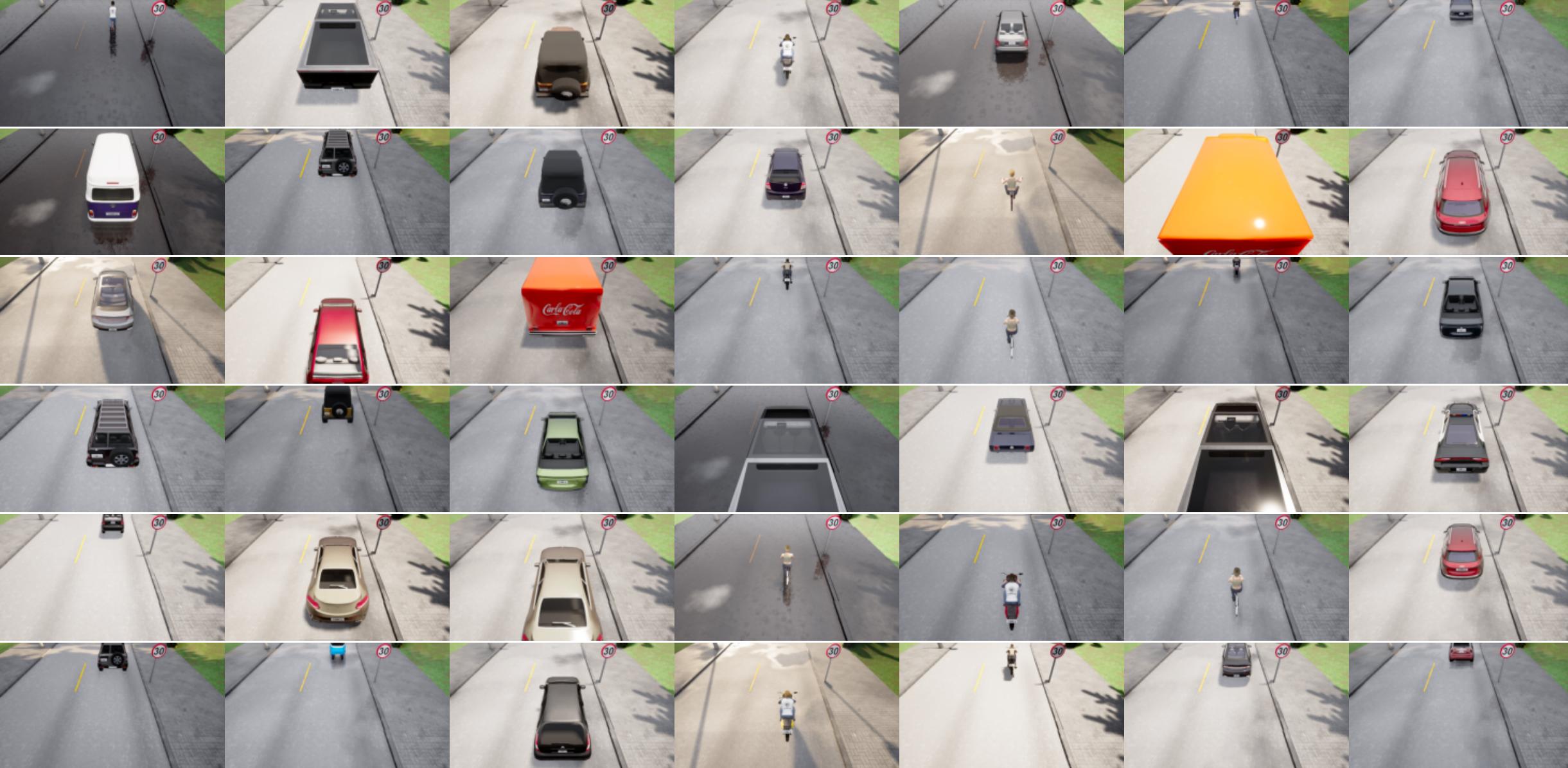}
    \caption{Examples of the synthetic dataset. Frame 30 of some of the episodes.}
    \label{fig:episodeImage}
    %\vspace{0mm}
\end{figure}

The generated synthetic vehicle speed detection dataset contains 610 episodes, where each episode is formed of a sequence of images which depend on the speed of the vehicle. Therefore, the dataset contains approximately 60,000 images labelled with vehicle speed, vehicle type and environmental conditions. Fig. \ref{fig:episodeImage} shows frame 30 of some of the synthetic episodes.

%The same distribution can be seen in Fig. \ref{fig:SpeedDistTrainValTest} for each of the different sets used for training \ref{fig:SpeedDistTrain}, validation \ref{fig:SpeedDistVal}, and test \ref{fig:SpeedDistTest}.

%\begin{figure}[ht]
%\centering
%    \begin{subfigure}[h]{.23\textwidth}
%        \centering
%        \includegraphics[width=\textwidth]{images/Train Speed Distribution.png}
%        \subcaption{Train set speed distribution.}\label{fig:SpeedDistTrain}
%    \end{subfigure}
%    \hfill
%    \begin{subfigure}[h]{.23\textwidth}
%        \centering
%        \includegraphics[width=\textwidth]{images/Validation Speed Distribution.png}
%        \subcaption{Validation set speed distribution.}\label{fig:SpeedDistVal}
%    \end{subfigure}
%    \hfill
%    \begin{subfigure}[h]{.23\textwidth}
%        \centering
%        \includegraphics[width=\textwidth]{images/Test Speed Distribution.png}
%        \subcaption{Test set speed distribution.}\label{fig:SpeedDistTest}
%    \end{subfigure}  
%    \caption{Distribution of speeds in the dataset (Train/Val/Test) in groups of 5 km/h}
%    \label{fig:SpeedDistTrainValTest}
%\end{figure}

%Using the CARLA simulator, it has been possible to generate, on a preliminary basis, a synthetic dataset with more than 600 episodes, with a high variability of vehicles, weather and lighting conditions in a relatively short time. The dataset can be easily extended by including a larger number of cameras with different points of view and new vehicles, which may favor the further development of speed detection systems based on deep learning. 

\subsection{Sequence-to-Sequence regression}

Vehicle speed detection is approached as a regression problem in which the model learns how to map the spatial and temporal relationships of a sequence of images to the corresponding vehicle speed. Two different models are adapted, implemented and validated, which will be described below. 

%In order to train a CNN capable of performing accurate speed detection, two different neural network models have been used, which will be described below. The first model is based on a 3D ResNet (see Fig. \ref{fig:scheme}(a)), and the second one is based on a combination of the VGG16 feature extraction network to which layers of a GRU network have been added to perform speed detection (see Fig. \ref{fig:scheme}(b)). 

\begin{figure}[t]
    \centering
    \includegraphics[width=\linewidth]{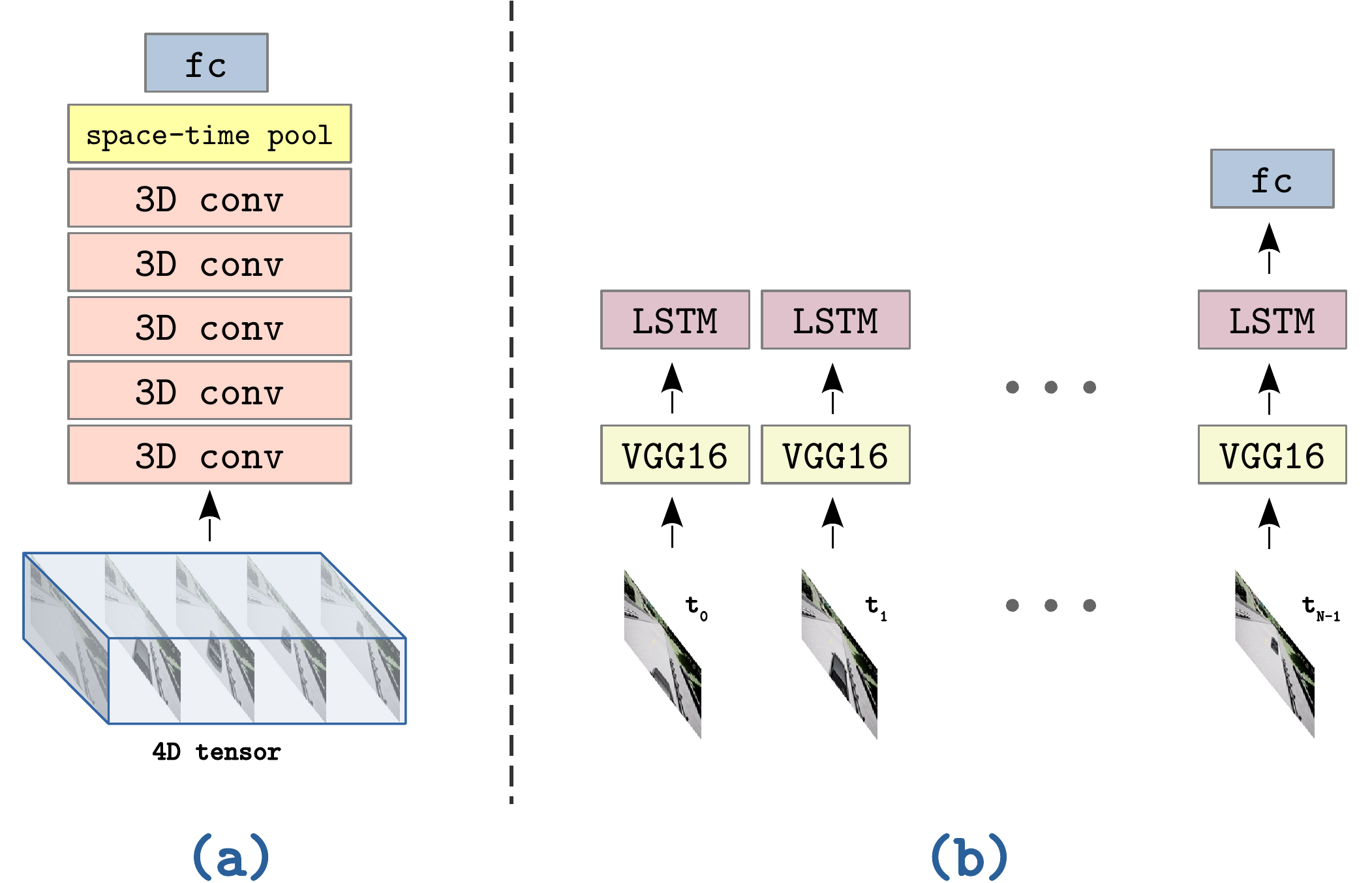}
    \caption{(a) 3D CNN (3D ResNet) and (b) CNN-RNN (VGG16-GRU).}
    \label{fig:scheme}
    %\vspace{0mm}
\end{figure}

\subsubsection{3D CNN}

The idea behind the 3D CNN model is to integrate spatial and temporal information into a 4D input tensor, and stack 3D convolutional layers with 3D convolutional filters that implicitly learn spatial and temporal features (see Fig. \ref{fig:scheme}(a)). In \cite{Ji2013}, the 3D CNN model was successfully applied for the first time to recognize human actions from video sequences, using a sliding window of 7 frames of size $60\times{}40$ centered on the current frame as inputs to the model. This work explains how to perform 3D convolutions, using 3D kernels that are applied to the cube formed by stacking multiple contiguous frames together. Thus, the feature maps in the convolutional layers are connected to multiple contiguous frames in the previous layer, thereby capturing motion information. 

%In order to train this type of networks, it is necessary to add the temporal dimension to the input data tensor of the network, which would have tensors $T\times{}W\times{}H\times{}3$ in the case of RGB images, and $T\times{}W\times{}H$ for grayscale images. In \cite{Ji2013} can be seen a 3D network of this type, used to recognize human actions from temporal sequences of images, giving the CNN as input a tensor of $7\times{}60\times{}40$. In addition, this work explains how to perform a 3D convolution, three-dimensional kernels are used and the operation is performed on the cube formed by the stacked images in time, obtaining a feature map for each set of frames selected by the size of the kernel, for example, if the kernel used has a temporal dimension of 3 frames, it will have a feature map for each set of 3 frames.

In our case, we propose to use 3D convolutional filters within the framework of residual learning. Following the approach proposed in \cite{Tran2018}, we apply a 3D ResNet18 architecture. Pre-trained models are usually obtained from human action classification datasets such as Kinetics-400 \cite{kay2017kinetics}, whose spatial and temporal features are very different from those that appear in our scenarios.  Therefore we perform training from scratch. The input is a 4D tensor of size $3\times{}N\times{}W\times{}H{}$, being $N=16$ the number of stacked frames which are obtained evenly spaced from a fixed time horizon. This implies that slower vehicles do not leave the camera's field of view, and for faster vehicles, the last images remain with the car having left the field of view (note that, in our preliminary study we consider that a sequence does not contain more than one vehicle).

%For this work, the network used was a 3D ResNet18, the same as shown in \cite{Tran2018}, and pre-trained on Kinetics-400\cite{kay2017kinetics}, a human action classification dataset. The input supplied to this network is a four-dimensional vector, being $C\times{}Frames\times{}W\times{}H{}$ . In this case, the last layer of the fully connected network has been modified so that it performs a regression, returning a single velocity value for each input sequence.

\subsubsection{CNN-RNN}

VGG16 \cite{simonyan2014_vgg16} has been chosen as the CNN feature extractor (i.e. freezing all layers). Using ImageNet \cite{Deng2009_imagenet} pre-trained weights, the output of the last convolutional layer has been used, with a dimension of $7 \times 7 \times 512$. This feature vector is flattened and used as input to the recurrent model, which is composed by a GRU \cite{cho2014_gru} layer with 50 units, an alternative to LSTM with similar performance and less computation requirements, and a dense layer with one output neuron. The number of timesteps is defined as $N=32$, corresponding to the features from evenly spaced images from a fixed time horizon (see Fig. \ref{fig:scheme}(b)).

%On the other side, there are recurrent networks such as LSTM, in this type of networks, each layer of the lstm is fed with a new input, as well as with its own output in a previous instant of time($t-1$), having this way a temporal perspective in the system.
%Combining this type of networks with a feature detector based on CNNs, each of the new inputs of the network would be the feature map of each of the consecutive frames of the sequence, building a CNN-LSTM as mentioned in \cite{Donahue2017}. In this last work, the network that they implement is able to recognize patterns in the image thanks to the CNN feature extractor phase, and at the same time, is able to recognize the actions performed in those images, thanks to the temporal information of the LSTM, giving a complete description of the picture.

\section{Experimental Evaluation}
\label{sec:results}

\subsection{Training parameters}
The total episodes recorded in the database are 610, split into 366 (60\%), 122 (20\%) and 122 (20\%) to generate the training, validation and test subsets, respectively. 

Adam optimizer and MSE loss are used in both models, with a learning rate of $3\times10^{-4}$ in the 3D ResNet case, and $10^{-4}$ in the CNN-GRU. Another difference between both models is in the batch size, and in the number of epochs, being 5 and 100 in the 3D model, and 3 and 150 in the RNN, respectively. Also, in the 3D-CNN model, early stopping is used, with a patience of 7, so the training end in epoch 25/100. To perform some regularization, the output  targets are normalized between -1 ($30 km/h$) and 1 ($100 km/h$).

The 3D CNN has a total of 33,166M trainable parameters, while the VGG16-GRU has 14,714M corresponding to the VGG16 pre-trained model and 3,771M to the trainable GRU architecture.

\subsection{Results}
\begin{figure}[ht]
    \def\svgwidth{\linewidth}
    \fontsize{6}{8}\selectfont
    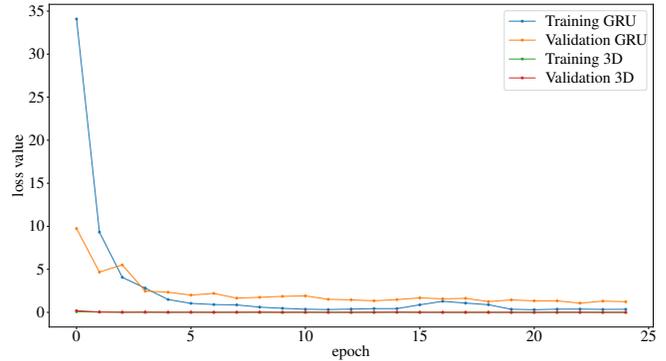
    \caption{Training and validation loss value temporal evolution.}
    \label{fig:lossEvaluation}
    %\vspace{0mm}
\end{figure}

The behaviour shown by both network architectures in the training and validation process is shown in Fig. \ref{fig:lossEvaluation}. The losses converge rapidly to values below 0.1. The general results are very satisfactory. On the one hand,  the architecture based on CNN-GRU presents an average speed error of $0.91m/s$ and the architecture based on 3D-CNN of $0.35m/s$. The two architectures have been compared using the same training, validation and test splits. 

% The simulated environmental conditions are ten and are identified by three values: angle of elevation of the sun over the day, percentage of precipitation and precipitation deposition on the surface. Although the simulator allows the modification of other parameters such as cloudiness or ambient fog, they have not been taken into account for simplification in the analysis of measurement errors depending on the climatic conditions. Thus, the solar elevation angle is discretised to two positions Midday ($75^{\circ}$) and Sunset ($15^{\circ}$), the precipitation percentage is discretised to values of (0, 15, 30 and 60), and the precipitation deposition percentage on the surface is discretised to (0, 50, 100). The label Midday\_15\_50 represents the solar angle, precipitation percentage, and precipitation deposition percentage on the surface. Ten environmental scenarios are defined and randomly selected for each simulation episode based on the above parameters. 

\begin{figure}[t]
    \def\svgwidth{\linewidth}
    \fontsize{5}{7}\selectfont
    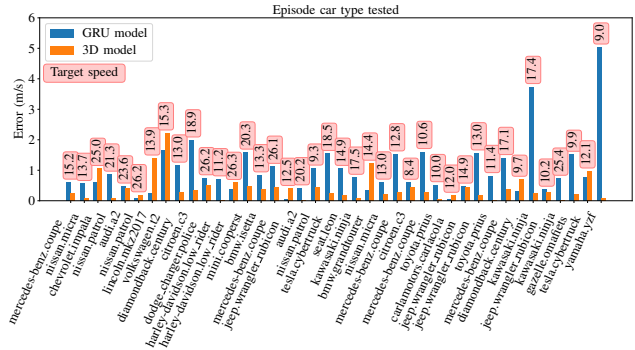

    \caption{Speed Error vs. Vehicle Type.}
    \label{fig:errorvsvehicle_0}
    %\vspace{0mm}
\end{figure}

\begin{figure}[ht]
    \def\svgwidth{\linewidth}
    \fontsize{5}{7}\selectfont
    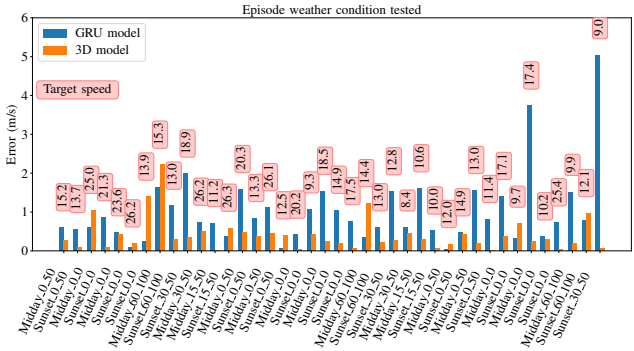
    \caption{Speed Error vs. Lighting and Weather conditions.}
    \label{fig:errorvsweather_0}
    %\vspace{0mm}
\end{figure}

Figs. \ref{fig:errorvsvehicle_0} and \ref{fig:errorvsweather_0} show the absolute error that the two architectures perform in the speed estimation of each test episode. Fig. \ref{fig:errorvsvehicle_0} combines information on the type of vehicle associated with the episode. With the CNN-GRU architecture, all the errors are below the threshold of $1.5m/s$, with two episodes with errors above $4m/s$ which correspond to two motorbike models (\emph{kawasaki.ninja} and \emph{yamaha.yzf}). These are small vehicles where the features generated may not be sufficiently representative.% where so we understand that the system has to improve the speed estimation for these vehicle models, extending the database in order to generalise the problem. 
%It could happen that for both episodes, their speed is at the limits of the simulated speeds or perhaps at a specific value. However, taking into account the episode's speed, it is $17m/s$ and $9m/s$, respectively (information in the pink table), which are estimated speeds with a low error in other episodes where the vehicle type is a car. Therefore, we will state that the GRU net is lacking in speed estimation for motorbikes. 
However, the 3D-CNN architecture net shows a stable behaviour, improving the estimation in all episodes concerning the CNN-GRU network, and mitigates the estimation problems found with the recurrent architecture because it obtains estimation errors in motorbikes close to $0.1m/s$.

On the other hand, Fig. \ref{fig:errorvsweather_0} integrates information on the environmental conditions associated with each episode. As in the previous case, the errors obtained by the CNN-GRU net are below the $1.5m/s$ threshold, and it is the two episodes studied above that present a higher error. If we analyse the environmental conditions associated with these episodes, \emph{Midday\_0\_0} and \emph{Sunset\_30\_50}, and compare them with the rest of the episodes, we can assert that the error is not related to the simulated lighting conditions. We infer that the error associated with these episodes come from the type of vehicle and the lack performance offered by the CNN-GRU net for these cases. On the other hand, and as in the previous analysis, we the 3D-CNN model provides accurate results robust to environmental conditions variability.

\begin{figure}[t]
    \def\svgwidth{\linewidth}
    \fontsize{5}{7}\selectfont
    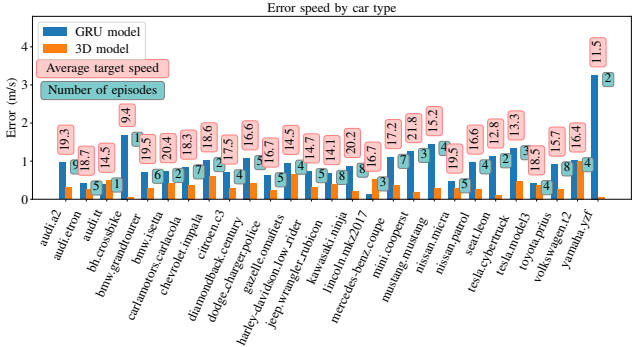
    \caption{Speed Error vs. Vehicle Type (general).}
    \label{fig:errorvsvehicle_1}
    %\vspace{0mm}
\end{figure}

Fig. \ref{fig:errorvsvehicle_1} shows the speed error function of simulated vehicle type, as well as the number of times the vehicle type appears in the test episodes (light green box) and average simulated speed for the vehicle in the set of episodes (light pink box) for both models. As in the previous case, in the CNN-GRU model, the error evaluated when the episodes include the aforementioned motorbikes stands out.
However, the two error values that rise above the rest have the common factor of simulating the slowest speeds in the set of episodes, i.e., $9.4m/s$ and $11.5m/s$. %Thus, we must include an analysis of how the simulated speed affects the error assesses. 
The minimum error is $0.5m/s$, which is linked to \emph{audi.etron}, \emph{nissan.micra} and \emph{tesla.model3} vehicle models, with simulated speeds between $18m/s$ and $19.5m/s$. The 3D-CNN model provides better results, with a maximum error of $1m/s$. % when the simulated vehicle is \emph{volkswagen.t2}. %However, we do not consider this error as a factor of increasing the number of times that the vehicle was simulated or the simulated average speed, depicting values of 8 and $15.7m/s$, respectively.

\begin{figure}[ht]
    \def\svgwidth{\linewidth}
    \fontsize{5}{7}\selectfont
    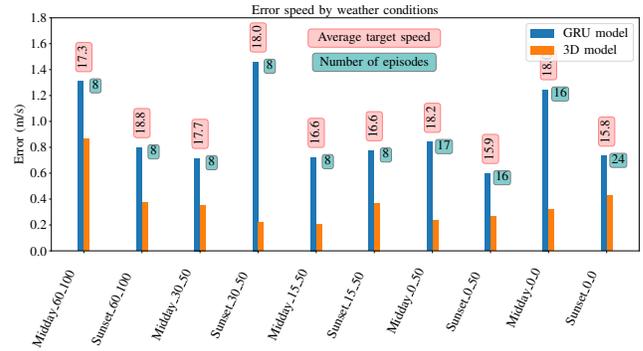
    \caption{Speed Error vs. Lighting and Weather conditions (general).}
    \label{fig:errorvsweather_1}
    %\vspace{0mm}
\end{figure}

On the other hand, Fig. \ref{fig:errorvsweather_1} depicts the absolute error in the speed estimation for the simulated lighting and weather conditions, as well as the number of times simulated (light green box) and the average simulated speed for the vehicle in the set of episodes (light pink box). The CNN-GRU model has a uniform behaviour for all simulated environmental conditions. The maximum error is $1.4m/s$, associated with Sunset, precipitation of 30\% and precipitation deposit of 50\%, which is linked with episodes where the motorbikes  were simulated. Again, the 3D-CNN model  provides more accurate and robust results in all the study cases, with errors lower than $0.8m/s$. 

\begin{figure}[t]

    \def\svgwidth{\linewidth}
    \fontsize{5}{7}\selectfont
    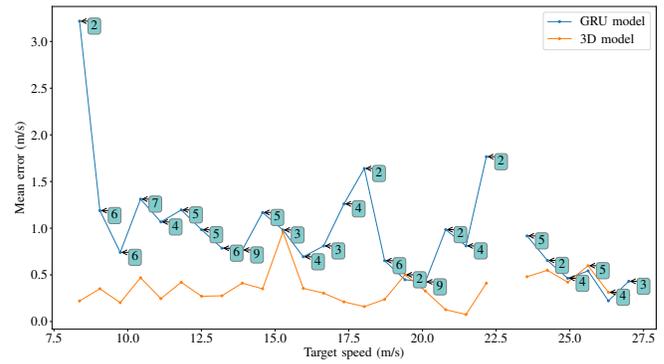
    \caption{Speed error with respect to the speed of the vehicles.}
    \label{fig:histogram}
\end{figure}

Finally, we analyze the mean absolute error as a function of the speed (Fig. \ref{fig:histogram}). Counter intuitively, in both cases, the estimation error does not seem to increase with vehicle speed. The number of samples is still relatively small, and the maximum speed is limited to 100 km/h. Although these preliminary results do not allow us to confirm whether this trend would occur at higher speeds, the fact that the error remains stable in the tests performed is very positive in view of an implementation of the system in real conditions. 

\begin{figure*}[ht]
    \centering
    \includegraphics[width=\linewidth]{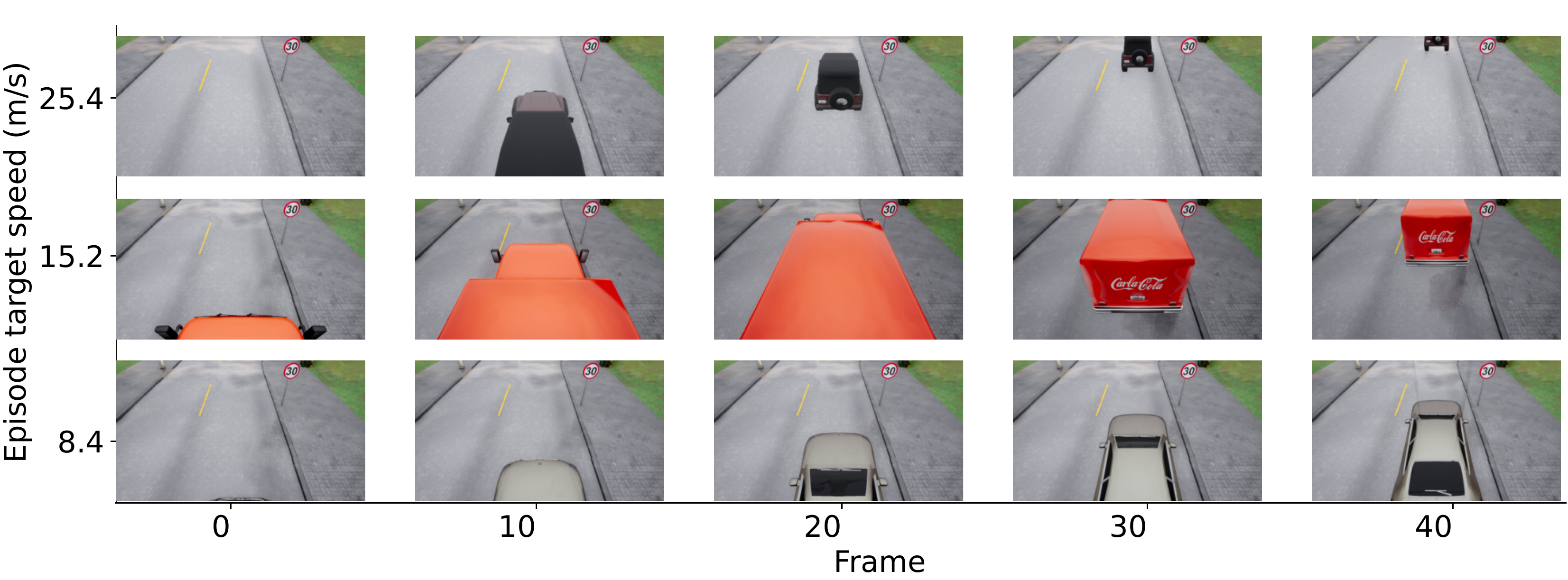}  
    \caption{Three episodes with three different speeds, with errors below $0.2m/s$. }
    \label{fig:results}
    %\vspace{0mm}
\end{figure*}

In Fig. \ref{fig:results} we present three episodes with three different speeds, in which the errors of both models were less than $0.2m/s$. 

%The recurrent model has a poorer performance in speed estimation when the simulated speed is slow, while the estimation improves as the speed increases. However, the 3D net has stable behaviour, with the estimated error not changing with the simulated speed. %The kinemometers base on recurrent neural networks, specifically those designed with 3D net, will be an alternative in coming steps. Technicals performed are below $1m/s$, in the case of the 3D net, below $0.4m/s$, fulfilling in both cases the requirements to gain a commercial system.

%\begin{figure}[ht]
%    \centering
%    \includegraphics[width=\linewidth]{images/ErrorVsWeather.eps}
%    \caption{Velocity histogram.}
%    \label{fig:histogram}
%    %\vspace{0mm}
%\end{figure}

\section{Conclusions}
 \label{sec:conclusion}
 This paper presents a learning-based accurate vehicle speed detection system using a new synthetic dataset, built by simulating a virtual camera in an autonomous driving simulator (CARLA). Data-driven approaches have so far not been robustly explored due to the difficulty in generating real datasets, with a sufficient number of vehicle sequences, with real speed values. The main contribution of our approach is based on the use, for the first time, of synthetic sequences for the generation of a vehicle speed detection system based on deep learning, with the necessary accuracy for its use as a speed enforcement system. A  synthetic dataset with more than 600 episodes has been generated, with a high variability of vehicles, weather and lighting conditions.  

Two different methods have been applied to address speed detection as a sequence-to-sequence regression problem using the synthetic sequences. First, a CNN-RNN model implemented with a VGG16 network to generate features from the images and a GRU layer to integrate temporal information which reports a MAE of $0.91 m/s$ (3.27 kmh). Second, a model based on 3D convolutions, implemented using a 3D ResNet, which provides a MAE of $0.35 m/s$ (1.26 kmh). These results, while still preliminary, are within the limits required by certification agencies, and validate our methodology as a basis for the development of accurate vision-based speed enforcement systems. 

However, this work has several limitations that need to be addressed in future work. For example, the range of speeds to be studied must be increased significantly. The dataset must be augmented with data from different cameras with variability in position and orientation, trying to address the problem independently of the camera perspective. Finally, the sim-to-real gap must be addressed and our methodology must be validated with real data sequences.

\section{Acknowledgements}

This work has been funded by research grant CLM18-PIC-051 (Community Region of Castilla la Mancha), and partially funded by research grants S2018/EMT-4362 SEGVAUTO 4.0-CM(Community  Region  of  Madrid) and DPI2017-90035-R (Spanish Ministry Science Innovation).

\bibliographystyle{IEEEtran}
\bibliography{references}
\end{document}